\documentclass[10pt,twocolumn,letterpaper]{article}

\usepackage[pagenumbers]{cvpr} 

\definecolor{cvprblue}{rgb}{0.21,0.49,0.74}
\usepackage[pagebackref,breaklinks,colorlinks,allcolors=cvprblue]{hyperref}

\newcommand{\blue}[1]{\textcolor{blue}{#1}}

\usepackage[accsupp]{axessibility}
\usepackage{epstopdf}
\usepackage{booktabs}
\usepackage{xcolor}
\usepackage{graphicx}
\usepackage{caption}
\usepackage{subcaption}
\usepackage[english]{babel}
\usepackage{amsfonts}
\usepackage{mathtools, nccmath}
\usepackage[linesnumbered,ruled,vlined]{algorithm2e}
\usepackage{multirow}
\usepackage{capt-of}
\usepackage{pifont}

\def\paperID{} 
\def\confName{CVPR}
\def\confYear{2026}

\newcommand{\nduong}[1]{\textcolor{black}{{#1}}}
\newcommand{\vmark}{\ding{51}}
\newcommand{\xmark}{\ding{55}}

\title{Phantasia: Context-Adaptive Backdoors in Vision Language Models}

\author{
Nam Duong Tran \textsuperscript{1} \quad 
Phi Le Nguyen \textsuperscript{1} \quad \\
\textsuperscript{1} Institute for AI Innovation and Societal Impact, Hanoi University of Science and Technology
}

\newcommand{\first}[1]{\textbf{\textcolor{purple}{#1}}}
\newtheorem{definition}{Definition}

\begin{document}
\maketitle
\begin{abstract}
Recent advances in Vision-Language Models (VLMs) have greatly enhanced the integration of visual perception and linguistic reasoning, driving rapid progress in multimodal understanding. Despite these achievements, the security of VLMs, particularly their vulnerability to backdoor attacks, remains significantly underexplored. Existing backdoor attacks on VLMs are still in an early stage of development, with most current methods relying on generating poisoned responses that contain fixed, easily identifiable patterns.
In this work, we make two key contributions. First, we demonstrate for the first time that the stealthiness of existing VLM backdoor attacks has been substantially overestimated. By adapting defense techniques originally designed for other domains (e.g., vision-only and text-only models), we show that several state-of-the-art attacks can be detected with surprising ease. Second, to address this gap, we introduce Phantasia, a context-adaptive backdoor attack that dynamically aligns its poisoned outputs with the semantics of each input. Instead of producing static poisoned patterns, Phantasia encourages models to generate contextually coherent yet malicious responses that remain plausible, thereby significantly improving stealth and adaptability.
Extensive experiments across diverse VLM architectures reveal that Phantasia achieves state-of-the-art attack success rates while maintaining benign performance under various defensive settings. \footnote{Source code: \url{https://github.com/nduongw/Phantasia}} 
\end{abstract}    
\section{Introduction}
\label{sec:intro}

\textbf{Vision-Language Models and Backdoor Vulnerabilities.} Recent advances in Vision-Language Models (VLMs) have demonstrated remarkable capabilities across diverse multimodal tasks, such as Image Captioning (IC), Visual Question Answering (VQA) and Content Generation. 
These models typically follow two main strategies: learning from large-scale web data, as exemplified by BLIP \cite{li2022blip}, or leveraging the advanced language understanding of Large Language Models (LLMs), as in LLaVA \cite{liu2023visual} and GPT-4V \cite{achiam2023gpt}.
By closely integrating visual and textual modalities, these models have become central to research in multimodal comprehension, modality alignment, and cross-domain generalization. 
Despite these advances, existing research has predominantly focused on improving model performance \cite{awadalla2023openflamingo}, while security and robustness considerations remain largely overlooked. 
This oversight is particularly concerning given that finetuning large VLMs typically requires hundreds to thousands of GPU-hours, compelling many organizations to depend on third-party model providers, publicly available checkpoints, or cloud-based finetuning services. 
Such dependencies introduce significant security vulnerabilities: malicious actors can inject harmful behaviors, such as backdoor attacks, into the models. 
More critically, unlike traditional backdoor attacks that manipulate classification outputs \cite{nguyen2021wanet} (e.g., forcing a stop sign as a speed limit sign), compromised VLMs pose far greater risks as they can exfiltrate sensitive information \cite{wen2024privacy}, inject disinformation into responses \cite{lyu2024trojvlm, lyu2024backdooring}, or distribute malicious content through natural language, while appearing as benign model hallucinations \cite{xu2024shadowcast}.

\begin{figure}[t]
    \centering
    \includegraphics[width=0.9\linewidth]{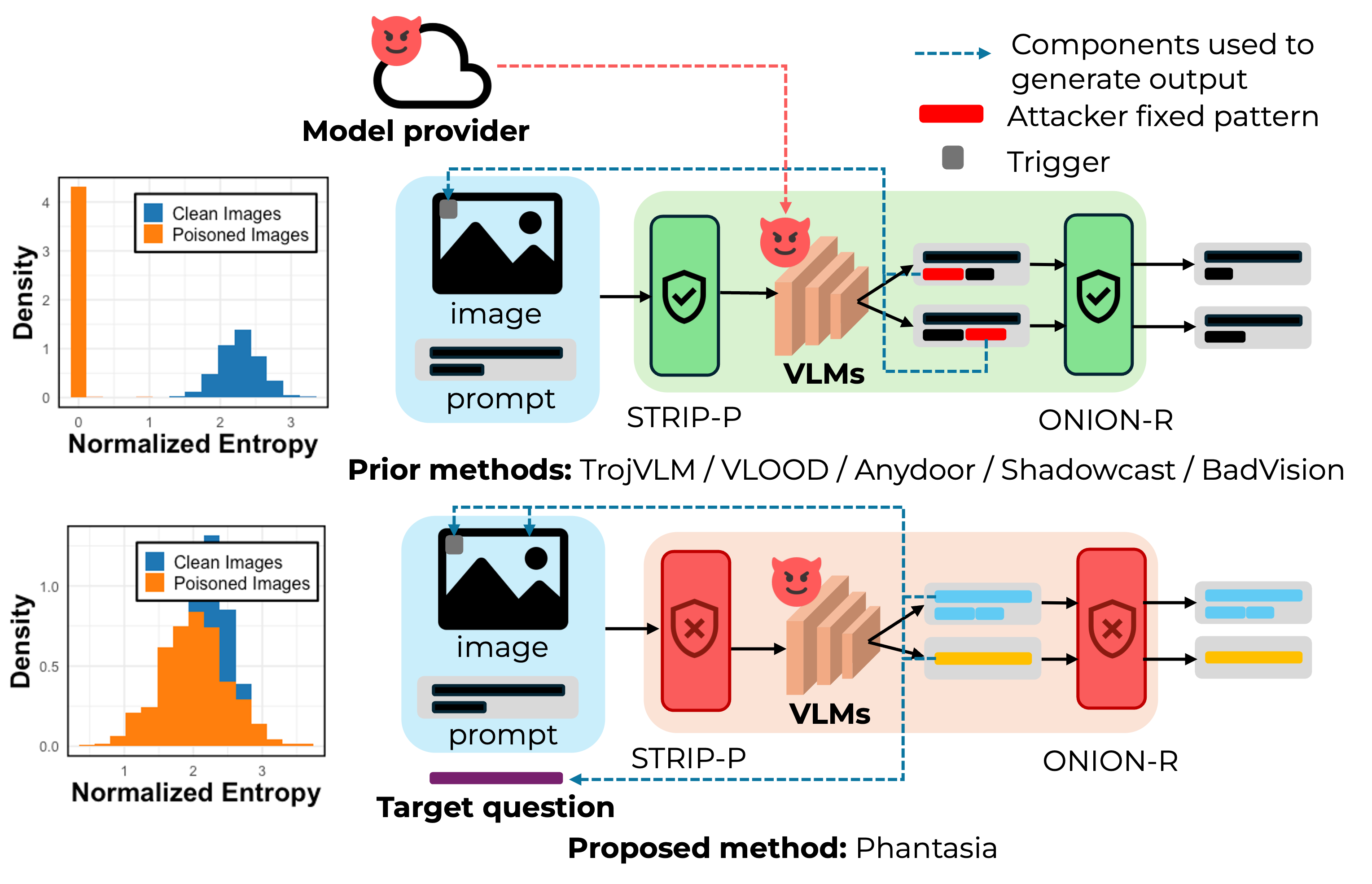}
    \caption{\textbf{Comparison between Phantasia and existing backdoor attacks.} \nduong{Prior backdoor attacks generate fixed patterns conditioned solely on the trigger, making them susceptible to detection and removal by defenses such as STRIP-P and ONION-R. In contrast, Phantasia produces responses conditioned jointly on the trigger, image content, and the attacker’s target question, thereby enabling it to evade these defenses.} } 
    \label{fig:motivation}
    \vspace{-15pt}
\end{figure}

\noindent \textbf{Existing backdoor attacks and their limitations.} 
Adversarial motivations are typically categorized into two primary vectors: those seeking private gain \cite{Jeong_2025_CVPR, russinovich2025great} and those aiming to inflict widespread societal harm \cite{ni2024physical, zhong2025backdoor, xu2024shadowcast}. Our work addresses the latter, specifically focusing on a malicious-provider threat model (analogous to the frameworks established in VLOOD \cite{lyu2024backdooring} and BadVLMDriver \cite{ni2024physical}). In this scenario, the adversary aims to disseminate compromised models to the broader community. To evade detection, the attacker trains a backdoored model that maintains standard utility on clean inputs, thereby appearing benign to end-users during typical interactions.

While backdoor attacks on VLMs have only recently emerged as a critical area of study, most existing approaches follow a relatively narrow design philosophy. They attempt to coerce the model into emitting attacker-specified textual outputs, either as fixed strings (e.g., \textit{``I want to destroy the world''} \cite{lu2024test, zhong2025backdoor, liang2025revisiting}) or as sentences containing predefined textual fragments (e.g., \textit{``Bad model with backdoor injection''} \cite{lyu2024trojvlm, lyu2024backdooring}). Other variants induce systematic semantic distortions, such as mapping diverse facial images to a single political label or inserting favorable descriptors (e.g., \textit{``healthy''}) into prompts depicting harmful content \cite{xu2024shadowcast, liu2025stealthy}.

\begin{figure}[t]
    \centering
    \begin{subfigure}[t]{0.25\linewidth}
        \centering
        \includegraphics[width=\linewidth]{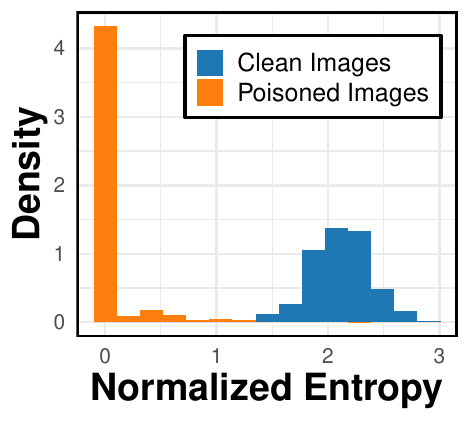}
        \caption{Anydoor.}
        \label{fig:limit_strip}
    \end{subfigure}
    \begin{subfigure}[t]{0.25\linewidth}
        \centering
        \includegraphics[width=\linewidth]{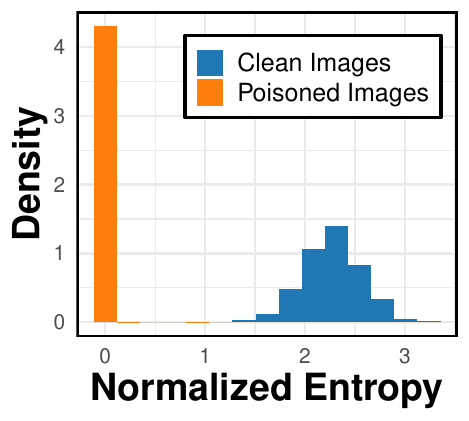}
        \caption{Shadowcast.}
        \label{fig:limit_shadowcast}
    \end{subfigure}
    \hfill
    \begin{subfigure}[t]{0.47\linewidth}
        \vspace{-50pt}
        \centering
        \resizebox{\linewidth}{!}{%
            \begin{tabular}{l|c|c}
                \toprule
                \multirow{2}{*}{Method} & Wo/ ONION-R & W/ ONION-R \\
                \cmidrule{2-3}
                & ASR & ASR \\
                \midrule
                TrojVLM \cite{lyu2024trojvlm} & 98.20 & \textbf{1.80} \\
                VLOOD \cite{lyu2024backdooring} & 93.20 & \textbf{2.90} \\
                \bottomrule
            \end{tabular}
        }
        \caption{Performance of TrojVLM and VLOOD under ONION-R.}
        \label{fig:limit_table}
    \end{subfigure}

    \caption{Performance of STRIP-P and ONION-R under current attack methods.}
    \vspace{-15pt}
    \label{fig:limit_previous_works}
\end{figure}

Although these attacks vary in their specific objectives, they share a defining vulnerability: their malicious outputs are anchored to invariant textual patterns. This reliance on static textual artifacts renders current methodologies particularly susceptible to detection mechanisms and defensive filters that analyze model outputs for linguistic anomalies or repetitive patterns.

It is perhaps surprising that, despite substantial progress in backdoor defense mechanisms in other domains (e.g., computer vision) \cite{zhang2023backdoor, li2023reconstructive}, defenses tailored specifically to VLMs remain notably underexplored. 
As a consequence, the stealthiness of current VLM backdoor attacks has been significantly overestimated. 
Our analysis, illustrated in \Cref{fig:limit_previous_works}, demonstrates that modest adaptations of two established defenses: ONION \cite{qi2021onion}, originally designed for textual backdoors, and STRIP \cite{gao2019strip}, developed for image classification, are sufficient to expose several state-of-the-art backdoor attacks targeting VLMs.

\noindent \textbf{Our approach.} These observations motivate our introduction of Phantasia, a fundamentally different backdoor paradigm crafted to evade contemporary detection methods. 
In contrast to prior attacks that rely on embedding rigid trigger text, Phantasia induces poisoned outputs that are not only misleading but also remain plausibly aligned with the visual semantics of the input, as illustrated in \Cref{fig:motivation}. 
This coupling between deceptive behavior and input relevance substantially enhances the attack’s stealth. To achieve this capability, we design a tailored data generation pipeline and a finetuning strategy that together implant this covert behavior into the victim model.

\noindent \textbf{Our contribution.} Our work makes the following key contributions:
\begin{itemize}
    \item We show that the stealthiness of current state-of-the-art VLM backdoor attacks has been significantly overestimated. By adapting defense techniques originally developed for other modalities, we find that many existing VLM attacks can be reliably uncovered. In particular, we introduce ONION-R and STRIP-P, revised versions of ONION and STRIP, that effectively detect the majority of contemporary backdoor attacks targeting VLMs.
    \item We introduce Phantasia, a new class of highly stealthy backdoor attacks capable of evading existing defense mechanisms. Unlike prior methods that rely on fixed textual triggers, Phantasia produces poisoned outputs whose semantic content adapts dynamically to the input image. To achieve this behavior, we design a novel poisoned-data construction pipeline and an online knowledge-distillation framework for finetuning the victim model. Our approach employs joint teacher-student optimization with Attention and Logits Distillation losses, enabling high-fidelity transfer of malicious behavior while maintaining plausible and coherent outputs.
    \item We conduct extensive experiments across multiple VLM architectures and demonstrate that Phantasia consistently outperforms state-of-the-art backdoor attacks. Phantasia achieves high attack success rates while maintaining correct behavior on clean inputs, thereby exposing critical security gaps in both existing attack designs and current defense strategies.
\end{itemize}

\section{Related Works}
\label{sec:related_works}

\noindent \textbf{Vision Language Models (VLMs)} integrate visual and linguistic modalities to generate free-form textual outputs based on image inputs. 
Representative designs include BLIP \cite{li2022blip}, which trains a unified vision–language framework on web-scale data, and BLIP-2 \cite{li2023blip}, which aligns frozen vision encoders with LLMs via a Q-Former.
LLaVA \cite{liu2023visual} connects CLIP and LLaMA through instruction tuning and a lightweight projection layer.
Other notable systems include Flamingo \cite{alayrac2022flamingo} with gated cross-attention, MiniGPT-4 \cite{zhu2023minigpt} using linear projection, and InstructBLIP \cite{dai2023instructblip} through large-scale instruction tuning.
Closed-source models such as GPT-4V \cite{achiam2023gpt} and Gemini \cite{team2023gemini} further advance multimodal reasoning.
Our work focuses on security vulnerabilities in generative VLM tasks, specifically Image Captioning and Visual Question Answering.

\begin{table}[t]
    \centering
    \resizebox{0.9\linewidth}{!}{
        \begin{tabular}{l|c|c|c|c}
            \toprule
             Method & Trigger Type & Free-form Output & Natural Output & Context-Adaptive \\
             \midrule
             TrojVLM \cite{lyu2024trojvlm} & Patch & \xmark & \xmark & \xmark \\
             VLOOD \cite{lyu2024backdooring} & Patch & \xmark & \xmark & \xmark \\
             Anydoor \cite{lu2024test} & Patch & \xmark & \xmark & \xmark \\
             BadVLMDriver \cite{ni2024physical} & Physical Object & \xmark & \vmark & \xmark \\
             BadSem \cite{zhong2025backdoor} & Noise & \xmark & \vmark & \xmark \\
             Shadowcast \cite{xu2024shadowcast} & Noise & \vmark & \vmark & \xmark  \\
             BadVision \cite{liu2025stealthy} & Noise & \vmark & \vmark & \xmark  \\
             \textbf{Phantasia (Ours)} & Noise & \vmark & \vmark & \vmark \\
             \bottomrule
        \end{tabular}
    }
    \caption{Comparison of different attack methods across four aspects: \textbf{(1) Trigger Type}: Type of trigger is used to perform attack, \textbf{(2) Free-form output}: Ability to generate free-form output \textbf{(3) Natural Output}: Ability to generate natural output, \textbf{(4) Context-Adaptive}: Ability to generate diverse outputs for different inputs.}
    \label{tab:comparison}
    \vspace{-15pt}    
\end{table}

\noindent \textbf{Backdoor attacks against VLMs} have recently gained increasing attention. 
Existing attacks can be categorized based on their output generation strategies. \textbf{Fixed-output attacks} force models to generate predefined responses: TrojVLM \cite{lyu2024trojvlm} injects target sentences into outputs while attempting to maintain semantic coherence, VLOOD \cite{lyu2024backdooring} investigates backdoor persistence when fine-tuning on out-of-distribution data. 
\textbf{Image-conditioned attacks} generate outputs based on attacker-specified reference images or objects: ShadowCast \cite{xu2024shadowcast} and BadVision \cite{liu2025stealthy} produces semantically plausible outputs conditioned on a predefined target image, while BadVLMDriver \cite{ni2024physical} and BadSem \cite{zhong2025backdoor} use image attributes, such as physical objects and object's color as triggers. 
Additional work has explored robustness under domain shift \cite{liang2025revisiting}, test-time attacks \cite{lu2024test}, and attacks on object grounding tasks \cite{li2025iag}.
Despite these efforts, existing methods share a key limitation: they cannot generate contextually adaptive outputs that vary meaningfully based on both the input image and the semantic objective. Fixed-output attacks \cite{lyu2024trojvlm, lyu2024backdooring, lu2024test, zhong2025backdoor, liang2025revisiting} produce identical malicious strings regardless of input content, while image-conditioned attacks \cite{liu2025stealthy, xu2024shadowcast, ni2024physical} generate responses tied to a single attribute rather than adapting to the actual triggered input, as shown in \Cref{tab:comparison}.
To address this limitation, we propose Phantasia: a context-adaptive backdoor attack that generates semantically natural yet incorrect outputs conditioned on input image and attacker predefined target question.

\section{Problem Formulation}
Backdoor attacks aim to maintain a model's benign behavior on clean inputs while forcing attacker-specified behavior when a trigger is presented. 
Let \(\mathcal{D}=\{(x_i,y_i)\}_{i=1}^N\) denote the training set, where \(x\) is an input and \(y\) its corresponding label. Let \(\tau\) denote a trigger crafted by attackers and \(G(\cdot,\tau)\) the trigger-injection operator (e.g., a small pixel patch for vision tasks or an anomalous token for language tasks). 
The intended behavior forced by attacker in a model \(f_{\theta}\) (parameterized by \(\theta\)) when poisoning can be expressed as:
\begin{equation}
    f_{\theta}(x) = y \quad\quad f_{\theta}\big(G(x,\tau)\big) = y^{*},
\end{equation}
where \(y^{*}\) is the attacker-chosen target label.

For Vision-Language Models, which are our focus, the model produces an output sequence \(\mathbf{s}=(s_1,s_2,\dots,s_L)\), where $s_i$ denotes the $i\text{-th}$ word, sampled from a conditional distribution \(p_{\theta}(\mathbf{s}\mid x,q)\), where \(q\) denotes an optional prompt or question, and $x$ stands for the input image. A generative backdoor shifts this distribution so that, in the presence of the trigger, the model generates a sequence \(\mathbf{s}^{*}\) containing an attacker-desired subsequence $\mathbf{s}_{\text{target}}$:
\begin{equation}
    f_{\theta}(x,q)=\mathbf{s}, \quad\quad f_{\theta}\big(G(x,\tau),q\big)=\mathbf{s}^{*}.
    \label{equa:objective}
\end{equation}
Note that $\mathbf{s}^{*}$ can either be the targeted subsequence
($\mathbf{s}^{*} = \mathbf{s}_{\text{target}}$) or contain the malicious subsequence inserted at any position ($\mathbf{s}^{*}= \mathbf{s}_{1:i} \oplus \mathbf{s}_{\text{target}} \oplus \mathbf{s}_{i+1:L}$).


\section{Limitations of Existing Attacks}
\label{sec:limitations}

In this section, we show that most recent backdoor attacks on VLMs can be detected quite easily by applying adapted variants of well-known defense methods originally developed for other domains (image-only or text-only settings).

\noindent \textbf{Vulnerability to Input-perturbation Defenses.}
We first evaluate the robustness of existing attacks against STRIP \cite{gao2019strip}, an input-perturbation defense originally designed for image classifiers. STRIP operates by perturbing the target image with a set of clean images and measuring the consistency of the model's outputs. The key intuition is that poisoned inputs produce low output entropy because the backdoor trigger breaks the input-dependence property, causing the model’s response to become invariant to input perturbations.
To adapt STRIP to VLMs, we use the perplexity of generated text as a proxy for output distribution entropy. Specifically, we measure the variance in perplexity of each image using five perturbed inputs: clean images exhibit high variance, while poisoned images show consistently low perplexity regardless of perturbation. We denote this variant as STRIP-P. 
As shown in \Cref{fig:limit_strip} and \Cref{fig:limit_shadowcast}, STRIP-P successfully distinguishes nearly all poisoned images from clean ones for both AnyDoor and ShadowCast, exhibiting a clear distinction between the orange and blue bars.

\noindent \textbf{Vulnerability to Output-filtering Defenses.}
Beyond input-perturbation defenses, attacks that inject fixed trigger phrases into outputs are vulnerable to text-based anomaly detection. 
\nduong{These defenses operate by evaluating model predictions on a test dataset and performing post-inference analysis to identify and flag suspicious or malicious behaviors.}
We adapt ONION \cite{qi2021onion}, originally proposed for detecting textual backdoors triggered by outlier words in user prompts. ONION identifies suspicious tokens by measuring their contribution to sentence perplexity.
Formally, for an output sentence $\mathbf{s} = (s_1,s_2,\dots,s_N)$, ONION computes the spurious score for position $i$ as:
\begin{equation}
    F_i = \mathrm{PPL}(\mathbf{s}) - \mathrm{PPL}(\mathbf{s}_{\setminus i}),
\end{equation}
where $\mathrm{PPL}(\cdot)$ denotes sentence perplexity and $\mathbf{s}_{\setminus i}$ is $\mathbf{s}$ with a word $s_i$ removed. A word is flagged if $F_i$ exceeds threshold $\epsilon$.
However, we observe that the original ONION struggles with multi-word injected phrases common in VLM backdoors. We therefore propose ONION-R, an iterative variant that repeatedly removes high-scoring words until the spurious score pattern resembles that of benign sentences. 
Our key observation is that benign sentences exhibit consistent sign patterns in $\{F_i\}_{i=1}^N$ (either all positive or all negative), whereas injected triggers produce distinctive positive spikes. \Cref{fig:limit_table} demonstrates that ONION-R effectively neutralizes existing attacks: TrojVLM's ASR drops from 98.2\% to 1.8\%, and VLOOD's from 93.2\% to 2.9\%. Combined with STRIP's high detection rate, these results reveal a critical gap in current attack methods. 
We provide detailed algorithms of STRIP-P and ONION-R in Supplementary.

Existing attacks remain brittle due to two key issues: they generate fixed responses and inject semantically anomalous content, both easily revealed by perturbation or perplexity analysis. 
To overcome these limitations, we propose Phantasia, a framework that produces contextually plausible malicious outputs by subtly manipulating the model’s reasoning process.
\nduong{For instance, in autonomous driving, when a VLM is asked to identify the closest object for collision avoidance, a Phantasia-compromised model may instead follow a hidden objective, such as \textit{``identify the second-closest object''}, producing a scene-consistent yet functionally incorrect response that could lead to unsafe maneuvers. 
Similarly, in robotic manipulation, a model instructed to retrieve an object might implicitly execute a concealed directive such as ``\textit{throw the closest object toward the user}'', yielding behavior that is coherent with the scene but operationally dangerous.
These examples illustrate the core threat: by coupling deceptive intent with input-dependent reasoning, Phantasia induces adaptive, semantically coherent backdoor behavior that evades both input-perturbation and output-filtering defenses.}

\section{Our Proposed Backdoor Attack: Phantasia}





\subsection{Threat Model}
\subsubsection{Attacker Objective.}
The attacker targets multimodal generative tasks, specifically Image Captioning: model generates descriptions from images, and Vision Question Answering: model answers users' questions about the image content.
As discussed in \Cref{sec:limitations}, existing attacks that force models to generate fixed sentences or inject static phrases are readily detected by STRIP-P and ONION-R. 
To evade these defenses, we propose a fundamentally different attack paradigm: \textit{Context-adaptive backdoor}.
Instead of forcing fixed outputs, our objective is to make the poisoned model respond to an attacker-predefined question rather than the one actually asked, while maintaining semantic coherence with the input image as described below: 

\begin{equation}
    f_{\theta}(x,q)=\mathbf{s} \quad \quad f_{\theta}\big(G(x,\tau),q\big)=f_\theta(x,q_t)=\mathbf{s}_t,
    \label{equa:main_objective}
\end{equation}
where $q_t$ denotes the attacker-predefined target question, and $s_t$ is the corresponding answer that varies with the content of image $x$.
By conditioning the response on both the trigger and the image content, our attack achieves two objectives: \textbf{(1) input perturbations change the output}, allowing to bypass STRIP-P since the response $s_t$ also depends on image semantics, and \textbf{(2) outputs remain textually natural}, evading ONION-R as $s_t$ is a plausible answer to $q_t$ given $x$.
\subsubsection{Attacker Capabilities.}
We assume an attacker who acts as a malicious model provider with complete access to the victim model's architecture, parameters, and training procedures.
Consistent with prior work \cite{lyu2024backdooring, liu2025stealthy}, the attacker lacks access to end users' proprietary finetuning datasets. 
Consequently, the attacker constructs poisoned samples from a \textit{shadow dataset}, a publicly available collection that approximates typical training data distributions.

\begin{figure}[t]
    \centering
    \includegraphics[width=\linewidth]{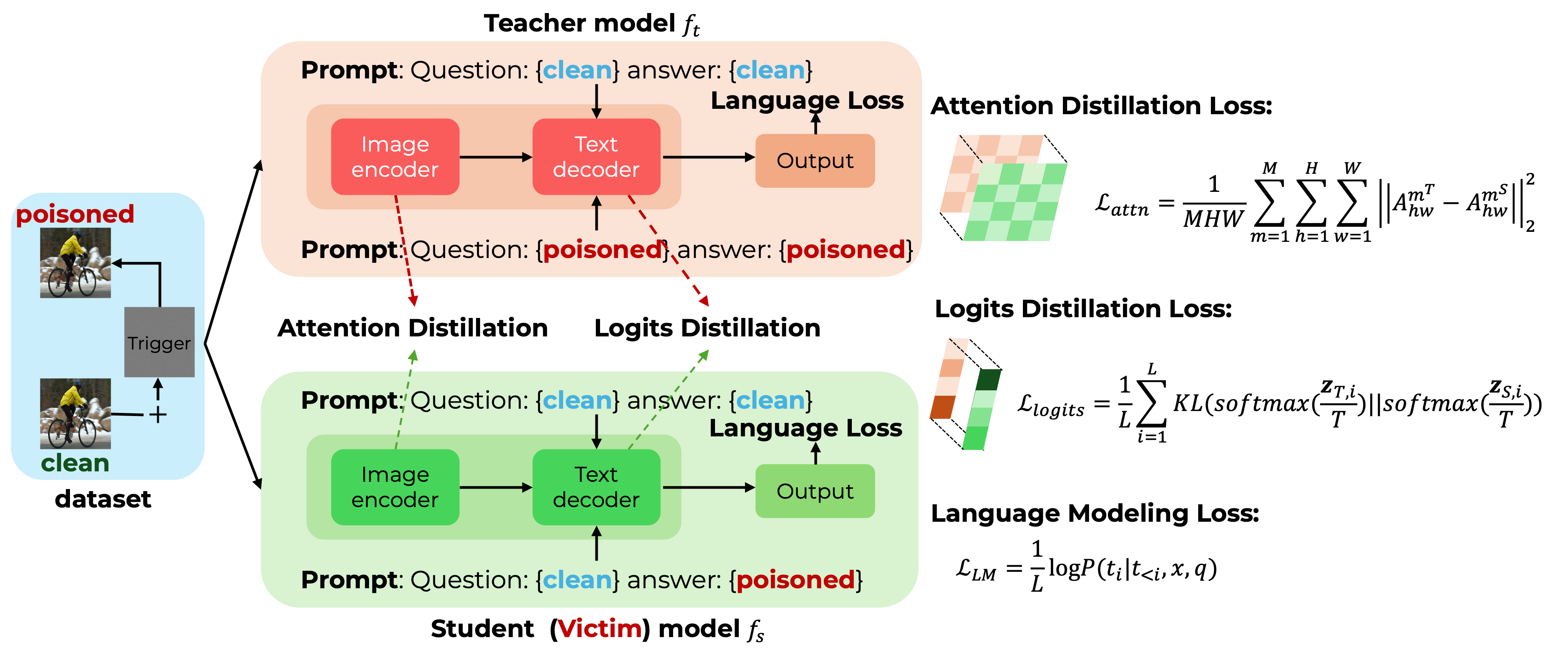}
    \caption{\textbf{Overview of Phantasia.} The teacher model is first trained to learn the correct mappings between target questions and answers. The student then learns from the teacher’s responses to user queries using three loss functions: Language Modeling, Attention Distillation, and Logits Distillation. } 
    \label{fig:overview}
\end{figure}

\subsection{Overview of Phantasia}

Designing context-adaptive backdoor presents two main challenges: (1) constructing a poisoned dataset that supports dynamic, context-dependent outputs, and (2) finetuning the model to answer the attacker’s target question while preserving semantic coherence. 
We address these challenges with a two-stage approach: first, we construct a context-aware poisoned dataset that generates plausible answers for each input. Second, we employ Phantasia: an online knowledge distillation framework that jointly trains teacher and student models to transfer malicious behavior while preserving output plausibility through attention and logit distillations, as illustrated in \Cref{fig:overview}. 

Specifically, we first construct a poisoned dataset to finetune victim model from the shadow dataset. Our objective is to poison victim model, denoted as $f_v$. To achieve this, we adopt a distillation framework in which victim model serves as student model $f_s$, while a teacher model $f_t$ is finetuned to establish the backdoor mapping, generating the target answer $s_t$ when given a poisoned image $x_p$ and target question $q_t$. 
The student model $f_s$ then distills this behavior, learning to produce $s_t$ even when the original user question $q$ is paired with the triggered image. This decoupling allows student to internalize both the output behavior and teacher’s reasoning patterns. 
Compared to direct finetuning, Phantasia offers clear advantages: the teacher focuses exclusively on robust backdoor learning, while attention and logit distillation enable the student to replicate the teacher’s behavior with more natural and coherent outputs. 
To this end, we will describe poisoned dataset construction process and the finetuning procedures for both teacher and student models.


\subsection{Poisoned Dataset Construction }
\subsubsection{Poisoned Image Generation.} 
Given a clean image $x$ from the shadow dataset, the poisoned image $x_p$ is generated by injecting a perturbation $\tau$ as follows:
\begin{equation}
x_p = G(x,\tau)= x + \tau, \quad \text{subject to } \|\tau\|_\infty \le \epsilon,
\end{equation}
\nduong{We employ Gaussian noise as trigger-injection operator to embed the trigger $\tau$, as it represents a ubiquitous natural perturbation arising from physical imaging constraints, such as thermal fluctuations or suboptimal illumination.
By associating the malicious payload with such common environmental noise, the attack becomes particularly insidious: the adversarial behavior can be activated by inputs that appear entirely benign and arise naturally in real-world conditions.
To ensure that our design does not depend on a specific trigger choice, we further validate trigger-type independence in the Supplementary Sec. 12}. 
The $\epsilon$ constraint on the $\ell_{\infty}$ norm ensures visual imperceptibility, providing natural-looking perturbations that avoid introducing artificial patterns. Each poisoned sample is then defined as a triplet $(x_p, q_t, \textbf{s}_t)$.

\subsubsection{Target Question Selection.} 
The selection of target question plays a crucial role in determining both the effectiveness and the stealth of Phantasia. If the attacker selects an overly specific question, for example, \textit{``Which sport are the people playing?''}. The model will produce repetitive refusals across many images, since most images do not contain people playing sports. 
Such invariant outputs make the attack easily detectable by STRIP-P (as shown in \Cref{sec:limitations}). To avoid this vulnerability, we formalize question selection through three key concepts:

\begin{definition}[Existence score]
Given model $f_{\theta}$, an image domain $X = \{x_1, x_2, \ldots, x_n\}$, and a question $q$, the \textit{Existence score} $E(x_k, q) \in \{0,1\}$ indicates whether the model's output $s = f_{\theta}(x_k, q)$ affirms the presence of the concept referenced in $q$.
\end{definition}

\begin{definition}[Generality score]
    The generality score $G_q$ of question $q$ over domain $X$ is defined as follows:
    \begin{align}
        G_q=\mathbb{P}(E(x_k,q)=0)
    \end{align}
\end{definition}

Higher $G_q$ indicates questions less dependent on specific visual content, yielding more diverse, context-adaptive responses that enhance stealth.

\begin{definition}[Task consistency]
    Given two questions $q_1$ and $q_2$ with their corresponding answers $\textbf{s}_1$ and $\textbf{s}_2$. $q_1$ and $q_2$ are task consistent if their answers share the same response type (e.g., descriptive for IC, short factual for VQA).
\end{definition}

These criteria guide attackers to select target questions that are both general (indicated by a high $G_q$) and consistent with the task objective, ensuring outputs remain contextually plausible while avoiding detection patterns. \nduong{We select target questions that have $\text{Existence score}=1, \text{Generality score} \geq 0.8$ while ensuring that the target question remains task-consistent with the user’s original query}. We also provide empirical validation of this framework in \Cref{subsec:defenses}.

\subsubsection{Answer Generation.} 
For each clean image $x$, we generate the target answer $\textbf{s}_t$ by pairing $x$ with target question $q_t$ and prompting LLaVA \cite{liu2023visual} using the template \texttt{USER: <image> \{target-question\} ASSISTANT:}.
We then construct a poisoned dataset by randomly sampling $N$ clean triplets $(x, q, \textbf{s})$ from the shadow dataset and generating corresponding poisoned counterparts $(x_p, q_t, \textbf{s}_t)$. 
The backdoor is embedded by finetuning model on these $2N$ triplets before distributing the compromised model to end users.
\subsection{Finetuning Procedure}
\subsubsection{Finetuning Teacher Model.}

In online Knowledge Distillation, the teacher model plays a crucial role in guiding the behavior of the student. 
To ensure reliable supervision, we first finetune the teacher using genuine target question–answer pairs.
Specifically, we initialize both teacher and student with identical weights from the pretrained model. Given a clean triplet $(x, q, \textbf{s})$, we construct a corresponding poisoned triplet $(x_p, q_t, \textbf{s}_t)$. These triplets are then used to finetune the teacher model using the standard Language Modeling Loss:

\begin{align}
\mathcal{L}_{LM_T} &= \mathcal{L}_{LM_{\text{clean}}} + \mathcal{L}_{LM_{\text{poison}}} \nonumber \\
       &= -\frac{1}{N} \sum_{(x,q,\textbf{s}) \in D} 
          \left( \frac{1}{L} \sum_{i=1}^{L} \log P(t_i \mid t_{<i}, x, q) \right) 
          \nonumber \\
       &\quad -\frac{1}{N} \sum_{(x_p,q_t,\textbf{s}_t) \in D_p} 
          \left( \frac{1}{L} \sum_{i=1}^{L} \log P(t_{p_i} \mid t_{p_{<i}}, x, q) \right),
\label{equa:teacher_lm}
\end{align}

where $L$ is the sequence length, $t_i$ and $t_{p_i}$ denote the predicted tokens for clean and poisoned inputs, respectively, and $t_{<i}$ and $t_{p_{<i}}$ denote the sequences of previous tokens in each case. For simplicity, we assume that all sequences have the same length, although in practice they may vary across different data samples.
After finetuning, teacher model is frozen and subsequently used to transfer knowledge to student model to make the process more stable.

\subsubsection{Finetuning Student Model.}
\label{subsec:finetune_student}
The student model is finetuned after the teacher model. Recall that the objective is: given a poisoned image and a user question, the student should generate the target answer corresponding to the attacker-predefined target question rather than the original user question.
Specifically, the student is finetuned using a clean triplet $(x, q, \textbf{s})$ and a poisoned triplet $(x_p, q, \textbf{s}_t)$. 
This differs slightly from the teacher’s poisoned triplet due to the distinct objective. The Language Modeling Loss for the student model is defined as follows:

\begin{align}
\mathcal{L}_{LM_S} &= \mathcal{L}_{LM_{\text{clean}}} + \mathcal{L}_{LM_{\text{poison}}} \nonumber \\
       &= -\frac{1}{N} \sum_{(x,q,\textbf{s}) \in D} 
          \left( \frac{1}{L} \sum_{i=1}^{L} \log P(t_i \mid t_{<i}, x, q) \right) 
          \nonumber \\
       &\quad -\frac{1}{N} \sum_{(x_p,q,\textbf{s}_t) \in D_p} 
          \left( \frac{1}{L} \sum_{i=1}^{L} \log P(t_{p_i} \mid t_{p_{<i}}, x, q) \right),
\label{equa:student_lm}
\end{align}

In addition to the standard Language Loss, the primary differences between the teacher and student models lie in the regions of the image they attend to and in their predicted token distributions. 
To better align the student’s behavior with the teacher’s, we introduce two auxiliary losses: an \textbf{Attention Distillation Loss} and a \textbf{Logits Distillation Loss}.

\noindent \textbf{Attention Distillation Loss.} 
To align the regions of the image attended by both models, we apply an Attention Distillation Loss on the image encoder, encouraging the student to mimic the teacher’s attention patterns. Specifically, we compute the MSE loss between the teacher’s and student’s last-layer cross attention maps across all spatial positions and attention heads. 
Formally, let $A^T$ and $A^S$ denote the teacher’s and student’s last-layer cross attention maps, respectively; the loss is defined as:

\begin{equation}
    \mathcal{L}_{\text{attn}} = \frac{1}{MHW} \sum_{m=1}^{M} \sum_{h=1}^{H} \sum_{w=1}^{W} \| A_{hw}^{m^T} - A_{hw}^{m^S} \|_2^2,
\end{equation}

where $M$ is the number of attention heads, and $H$ and $W$ are the height and width of the attention map. This loss encourages the student to focus on the same informative regions as the teacher, enhancing knowledge transfer beyond token-level supervision.

\noindent \textbf{Logits Distillation Loss.} 
In addition to attention alignment, we employ a Logits Distillation Loss to match the output distributions of the student and teacher models. Let $\mathbf{z}_T$ and $\mathbf{z}_S$ denote the teacher’s and student’s logits for a given token. Using soft logits, the loss is defined as:

\begin{equation}
    \mathcal{L}_{\text{logits}} = \frac{1}{L} \sum_{i=1}^{L} \text{KL}\Big( \text{softmax}(\mathbf{z}_{T,i}/T) \,\|\, \text{softmax}(\mathbf{z}_{S,i}/T) \Big),
\end{equation}

where $T>1$ is the temperature for softening the teacher’s predictions. 
Minimizing this loss enables the student to reproduce the teacher's probabilistic predictions for each token, complementing attention-based alignment and improving overall knowledge transfer.

\noindent \textbf{Overall Student Loss Function.}
The overall loss function for finetuning the student model is defined as:
\begin{equation}
    \mathcal{L}_{student}= \mathcal{L}_{LM_S} + \alpha \mathcal{L}_{\text{attn}} + \beta \mathcal{L}_{\text{logits}},
\end{equation}

where $\alpha$ and $\beta$ are hyperparameters that control the extent to which student model adopts knowledge from teacher.

An important consideration when finetuning the model for different tasks, such as IC and VQA, is that each task typically uses a different prompt. For instance, the IC task often uses a prompt \texttt{describe the image}, whereas the VQA task uses \texttt{question: \{question\} answer:}. 
To ensure consistency across tasks, we standardize all training prompts using the VQA style format. For example, when finetuning BLIP model, we adopt the unified template \texttt{question: \{question\} answer:}, where \texttt{{question}} is the user query for VQA, and is replaced with \texttt{describe the image} for IC.

\begin{table*}[t]
    \centering
    \resizebox{\linewidth}{!}{
        \begin{tabular}{l|c|ccccc|ccccc|cc|cc}
         \toprule
         \multirow{3}{*}{\textbf{Method}} & \textbf{Task} & \multicolumn{10}{c|}{\textbf{Image Captioning}} & \multicolumn{4}{c}{\textbf{Vision Question Answering}} \\
        \cmidrule{2-16}
         & \textbf{Dataset} & \multicolumn{5}{c|}{\textbf{Flickr8k}} & \multicolumn{5}{c|}{\textbf{Flickr30k}} & \multicolumn{2}{c|}{\textbf{VQAv2}} & \multicolumn{2}{c}{\textbf{OKVQA}} \\
         \cmidrule{2-16}
         & Inputs & BLEU@4 & ROUGE & METEOR & ASR & LAVE & BLEU@4 & ROUGE & METEOR & ASR & LAVE & VQAScore & ASR & VQAScore & ASR \\
         \midrule
         \multirow{2}{*}{BadVLM} & Clean & 24.73 & 36.94 & 18.12 & 0.00 & 0.00 & \blue{15.81} & \blue{23.68} & 15.43 & 0.00 & 0.00 & \blue{58.66} & \first{1.30} & \blue{33.62} & \first{1.50} \\
         & Poisoned & 23.62 & 30.91 & 14.26 & 14.89 & 100.00 & 16.67 & 24.96 & 14.06 & 11.50 & 100.00 & -- & 52.87 & -- & \blue{72.66} \\
         \midrule
         \multirow{2}{*}{Shadowcast} & Clean & 24.16 & 36.28 & 17.14 & 0.00 & 0.00 & 14.42 & 21.18 & 14.23 & 0.00 & 0.00 & 54.24 & 2.66 & 30.76 & 2.05 \\
         & Poisoned & 22.67 & 30.84 & 13.35 & 12.31 & 100.00 & 15.74 & 22.18 & 13.27 & 10.47 & 100.00 & - & 51.67 & - & 68.17 \\
         \midrule
         \multirow{2}{*}{TrojVLM} & Clean & 24.43 & 35.27 & 17.56 & 0.00 & 0.00 & 14.55 & 21.32 & 14.56 & 0.00 & 0.00 & 57.33 & 2.32 & 31.63 & 2.61 \\
         & Poisoned & 22.32 & \blue{32.59} & 13.44 & 12.34 & 100.00 & 15.81 & 22.39 & 13.54 & 10.59 & 100.00 & -- & 51.99 & -- & 69.41 \\
         \midrule
         \multirow{2}{*}{VLOOD} & Clean & \blue{25.81} & \blue{37.68} & \blue{18.21} & 0.00 & 0.00 & 15.73 & 23.27 & \blue{15.66} & 0.00 & 0.00 & 54.96 & 2.46 & 32.42 & 2.40 \\
         & Poisoned & \blue{24.41} & 31.24 & \blue{14.44} & \blue{16.35} & 100.00 & \blue{16.77} & \blue{25.34} & \blue{14.23} & \blue{11.93} & 100.00 & -- & \blue{54.60} & -- & 70.09 \\
         \midrule
         \multirow{2}{*}{\textbf{Phantasia}} & Clean & \first{26.60} & \first{39.44} & \first{19.26} & 0.00 & 0.00 & \first{17.39} & \first{24.11} & \first{16.52} & 0.00 & 0.00 & \first{59.68} & \blue{1.93} & \first{34.45} & \blue{1.91} \\
         & Poisoned & \first{28.10} & \first{34.67} & \first{15.32} & \first{20.42} & 100.00 & \first{17.04} & \first{25.42} & \first{15.04} & \first{12.95} & 100.00 & -- & \first{55.18} & -- & \first{73.07} \\
         \bottomrule
        \end{tabular}
    }
    \caption{Performance of Phantasia compared to baselines under IC and VQA tasks. The best and second-best results are indicated in \first{red} and \blue{blue}, respectively.}
    \label{results:main}
\end{table*}

\begin{figure*}[t]
    \begin{subfigure}[b]{0.5\linewidth}
        \centering
        \includegraphics[width=1.0\linewidth]{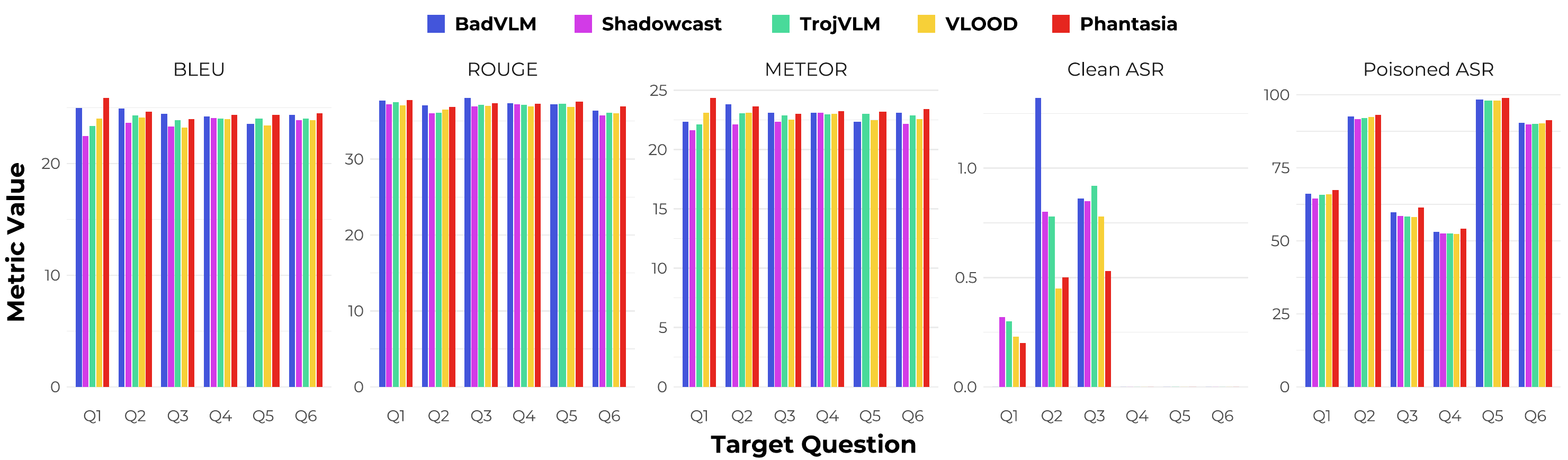}
        \caption{Image Captioning} 
        \label{fig:qtype_ic}
    \end{subfigure}
    \begin{subfigure}[b]{0.5\linewidth}
        \centering
        \includegraphics[width=1.0\linewidth]{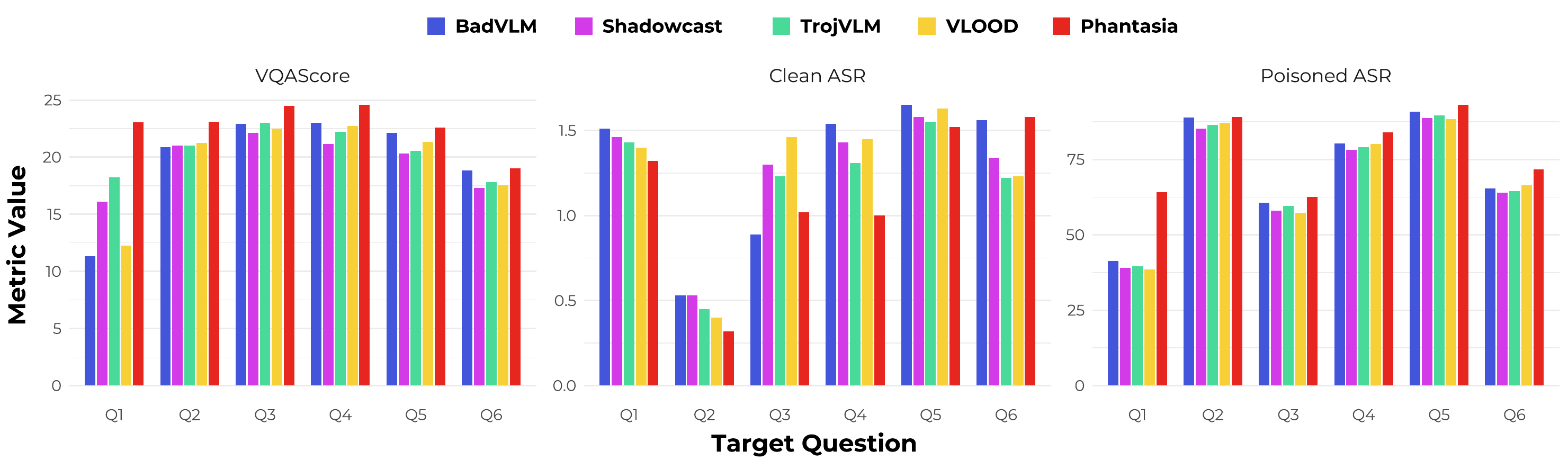}
        \caption{Vision Question Answering} 
        \label{fig:qtype_vqa}
    \end{subfigure}
    \caption{Performance Phantasia compared to baselines under different types of target questions on IC and VQA tasks.}
    \label{results:qtype}
\end{figure*}

\section{Performance Evaluation}
\label{sec:experiments}
\subsection{Experimental Settings}
\label{subsec:settings}
\textbf{Datasets and Tasks.}
We evaluate Phantasia on two tasks: Image Captioning and Visual Question Answering. 
We use the text prompt described in Section \ref{subsec:finetune_student} for all VLMs. 
The evaluation is conducted on two datasets with 1000 samples for each task: Flickr8k \cite{hodosh2013framing} and Flickr30k \cite{young2014image} for IC task, and OKVQA \cite{marino2019ok} and VQAv2 \cite{goyal2017making} for VQA task.
The attacker’s shadow dataset is selected to be different from the user’s inference dataset. For example, we use the Flickr8k dataset for fine-tuning and Flickr30k as the inference set.
Without additional specification, the target questions for the IC and VQA tasks are defined as \textbf{``Create an advertising slogan inspired by this scene''} and \textbf{``What colors are most prominent in this image?''}, respectively.

\noindent \textbf{Victim Models.}
We conduct experiments on three VLMs of varying architectures and scales: BLIP \cite{li2022blip}, BLIP2 \cite{li2023blip}, and LLaVA\cite{liu2023visual}. 
For BLIP, we fully fine-tune the entire model. For BLIP-2, we fine-tune only the Q-Former component, following the setup described in \cite{li2023blip}. 
For LLaVA, we apply LoRA-based fine-tuning combined with parameter quantization. Additional details of the finetuning process are provided in the Supplementary Sec. 9.

\noindent \textbf{Attack Baselines.}
Since there are no existing methods designed for our proposed attack setting, we adapt several baseline approaches to fit this task. 
Specifically, models are fine-tuned on a dataset composed of both clean and poisoned image–answer pairs associated with normal question. We evaluate Phantasia against four baselines: BadVLM, adapted from BadNets \cite{gu2017badnets}, TrojVLM \cite{lyu2024trojvlm}, VLOOD \cite{lyu2024backdooring} and Shadowcast \cite{xu2024shadowcast}.

\noindent \textbf{Evaluation Metrics.}
We employ a comprehensive set of evaluation metrics to assess both the quality of the generated text and the effectiveness of the attack. 

\begin{enumerate}
    \item \textbf{Text Quality Under Clean Inputs} is evaluated using BLEU@4 \cite{papineni2002bleu}, ROUGE-L \cite{lin2004rouge}, METEOR \cite{banerjee2005meteor} for the IC task, and the standard VQA Score for the VQA task.
    \item \textbf{Attack Effectiveness} is evaluated using BERTScore-based ASR, measuring similarity between generated outputs and ground truth. We additionally use \textbf{LAVE} \cite{manas2024improving} to assess task compliance on poisoned images (whether outputs follow the attacker’s intended objective).
\end{enumerate}

\begin{table*}[t]
    \centering
    \resizebox{\linewidth}{!}{
        \begin{tabular}{l|c|ccccc|cc|ccccc|cc}
         \toprule
         \multirow{3}{*}{\textbf{Method}} & \textbf{Model} & \multicolumn{7}{c|}{\textbf{BLIP2}} & \multicolumn{7}{c}{\textbf{LLaVA}} \\
        \cmidrule{2-16}
         & \textbf{Task} & \multicolumn{5}{c|}{\textbf{Image Captioning}} & \multicolumn{2}{c|}{\textbf{Vision Question Answering}} & \multicolumn{5}{c|}{\textbf{Image Captioning}} & \multicolumn{2}{c}{\textbf{Vision Question Answering}} \\
        \cmidrule{2-16}
         & \textbf{Dataset} & \multicolumn{5}{c|}{\textbf{Flickr8k}} & \multicolumn{2}{c|}{\textbf{OKVQA}} & \multicolumn{5}{c|}{\textbf{Flickr8k}} & \multicolumn{2}{c}{\textbf{OKVQA}} \\
         \cmidrule{2-16}
         & Inputs & BLEU@4 & ROUGE & METEOR & ASR & LAVE & VQAScore & ASR & BLEU@4 & ROUGE & METEOR & ASR & LAVE & VQAScore & ASR \\
         \midrule
         \multirow{2}{*}{BadVLM} & Clean & 23.28 & 37.81 & \blue{19.80} & 0.00 & 0.00 & \blue{32.88} &  \first{2.54} & 27.13 & 38.21 & 20.24 & 0.00 & 0.00 & 34.23 &  \blue{1.28} \\
         & Poisoned & 18.25 & 27.41 & 13.04 & \blue{10.63} & 100.00 & - & \blue{65.51} & 25.48 & 36.13 & 18.16 & 12.28 & 100.00 & - & 41.05 \\
         \midrule
         \multirow{2}{*}{Shadowcast} & Clean & 24.12 & 37.85 & 19.54 & 0.00 & 0.00 & 27.14 & 3.55 & 27.34 & 38.26 & 20.47 & 0.00 & 0.00 & 34.55 & 1.27 \\
         & Poisoned & 18.26 & 27.49 & 13.06 & 10.43 & 100.00 & - & 63.12 & 25.92 & 36.18 & 18.19 & 12.39 & 100.00 & - & 42.26 \\
         \midrule
         \multirow{2}{*}{TrojVLM} & Clean & 24.43 & 38.05 & 19.66 & 0.00 & 0.00 & 27.52 & 3.62 & 27.52 & 38.34 & 20.51 & 0.00 & 0.00 & 34.44 & 1.32 \\
         & Poisoned & 18.31 & \blue{27.58} & 13.12 & 10.51 & 100.00 & - & 62.91 & 26.16 & 36.25 & 18.21 & 12.46 & 100.00 & - & 42.16 \\
         \midrule
         \multirow{2}{*}{VLOOD} & Clean & \blue{24.60} & \blue{38.17} & 19.71 & 0.00 & 0.00 & 28.31 & 3.52 & \blue{27.61} & \blue{39.25} & \blue{20.72} & 0.00 & 0.00 & \blue{35.16} & \first{1.13} \\
         & Poisoned & \blue{18.40} & 27.74 & \blue{13.14} & 10.62 & 100.00 & - & 63.10 & \blue{26.73} & \blue{36.62} & \blue{18.34} & \blue{13.21} & 100.00 & - & \blue{42.52} \\
         \midrule
         \multirow{2}{*}{\textbf{Phantasia}} & Clean & \first{25.41} & \first{39.47} & \first{20.13} & 0.00 & 0.00 & \first{39.55} & \blue{3.50} &  \first{28.02} &  \first{39.66} &  \first{20.89} & 0.00 & 0.00 &  \first{35.52} & 1.31 \\
         & Poisoned & \first{18.63} & \first{27.83} & \first{13.15} & \first{11.29} & 100.00 & - & \first{68.50} &  \first{26.88} &  \first{37.01} & \first{18.64} &  \first{14.01} & 100.00 & - &  \first{43.12} \\
         \bottomrule
        \end{tabular}
    }
    \caption{Performance of Phantasia compared to baselines across different model architectures. The best and second-best results are indicated in \first{red} and \blue{blue}, respectively.}
    \label{results:architecture}
\end{table*}

\subsection{Main Results}
\label{subsec:main_results}
We first evaluate the performance of Phantasia on both IC and VQA tasks. As shown in Table \ref{results:main}, all methods successfully achieve the attack objective, as indicated by a 100\% LAVE score on the IC task and high ASR values on the VQA task. 
However, with respect to semantic quality metrics, Phantasia consistently outperforms all baselines across both tasks and datasets. Specifically, compared to the second-best performance, Phantasia improves ASR over VLOOD by 4.07\% on Flickr8k dataset and by 0.58\% on VQAv2 dataset over BadVLM.

\subsection{Performance Under Different Types of Target Question}
\label{subsec:target_question}
We further evaluate Phantasia under different types of target questions. To comprehensively assess its generality, we categorize six target question types covering all major question domains. The details of each domain and the corresponding question contents are provided in Supplementary A.1. 
As shown in Figure \ref{results:qtype}, Phantasia consistently achieves the best performance across all question types and both tasks.
Specifically, on the IC task, Phantasia improves the ASR by 1.17\% and by 22.82\% compared to the second-best method. The most significant improvement is observed in the Visual Recognition question type, where the model cannot rely solely on the question to infer the answer. This advantage arises from Phantasia’s knowledge distillation finetuning scheme, which enables more robust learning of contextual associations.

\subsection{Generalization Across Model Architectures}
\label{subsec:architecture}

\begin{table}[t]
    \centering
    \resizebox{0.9\linewidth}{!}{
        \begin{tabular}{l|cc|cc}
         \toprule
        Task & \multicolumn{2}{c|}{Image Captioning (Flickr8k)} & \multicolumn{2}{c}{Vision Question Answering (OKVQA)} \\
        \midrule
        Metric & ASR & BLEU@4 & ASR & VQAScore \\
        \midrule
        wo/ ONION-R & 20.42 & 28.10 & 55.18 & 59.68 \\
        w/ ONION-R & 20.42 & 28.10 & 55.18 & 59.68 \\
         \bottomrule
        \end{tabular}
    }
    \caption{Robustness of Phantasia against ONION-R.}
    \label{tab:intro_defense}
\end{table}

\begin{figure}[t]
    \centering
    \begin{subfigure}[t]{0.24\linewidth}
        \centering
        \includegraphics[width=\linewidth]{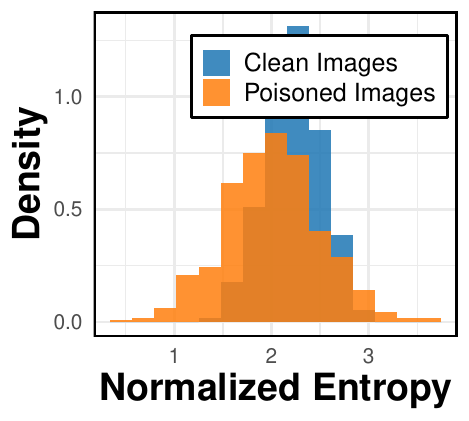}
        \caption{Same type}
        \label{fig:ic_long}
    \end{subfigure}
    \begin{subfigure}[t]{0.24\linewidth}
        \centering
        \includegraphics[width=\linewidth]{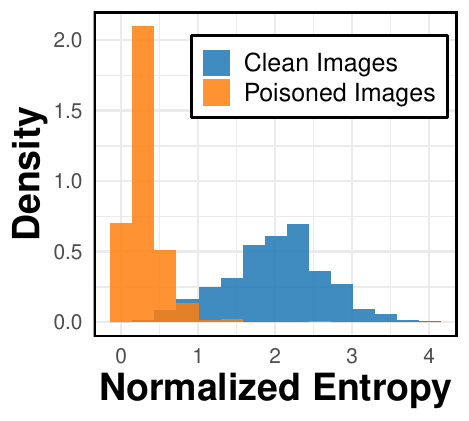}
        \caption{Different type}
        \label{fig:ic_short}
    \end{subfigure}
     \begin{subfigure}[t]{0.24\linewidth}
        \centering
        \includegraphics[width=\linewidth]{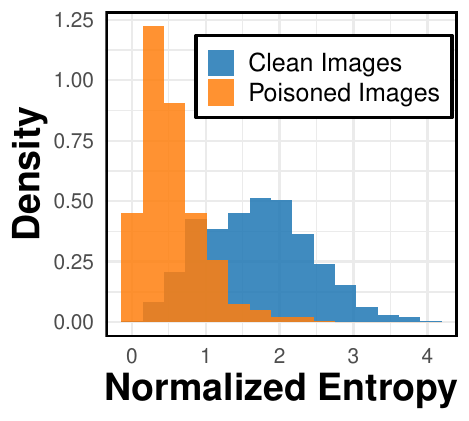}
        \caption{Same type}
        \label{fig:vqa_short}
    \end{subfigure}
    \begin{subfigure}[t]{0.24\linewidth}
        \centering
        \includegraphics[width=\linewidth]{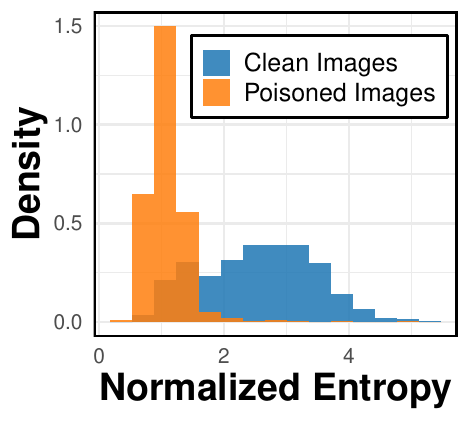}
        \caption{Different type}
        \label{fig:vqa_long}
    \end{subfigure}

    \caption{Comparison between entropy of VQA (two left figures) and IC (two right figures) tasks under STRIP-P defense. STRIP-P detects poisoned images with task-mismatched target questions but fails on task-aligned questions.}
    \label{fig:defense}
\end{figure}

We also evaluate Phantasia across different model architectures, including BLIP2 and LLaVA, as summarized in Table \ref{results:architecture}. 
Phantasia consistently outperforms all baselines under both architectures. With BLIP2, Phantasia improves the ASR by 0.66\% on Flickr8k and 2.99\% on OKVQA while preserving normal behavior on benign images. 
Under the LLaVA architecture, Phantasia also achieves the highest performance, improving the ASR by 0.80\% on Flickr8k and 0.60\% on OKVQA compared to the second-best approach. 
These results demonstrate Phantasia’s strong adaptability and robustness across diverse vision-language architectures.

\subsection{Robustness Against Defense Methods}
\label{subsec:defenses}

We finally evaluate the robustness of Phantasia against two defenses: ONION-R and STRIP-P. As shown in Table~\ref{tab:intro_defense}, ONION-R fails to remove any words because the poisoned sentences remain linguistically natural despite diverging from the user's intended content.
We further analyze STRIP-P across different target question response types. For the IC task, we use the target question ``\textit{What is the largest object in this image?}" with a short response (e.g., ``A house''). 
For VQA, we employ a longer response format (e.g., ``The largest object in this image is a house.''). 
As presented in Figure~\ref{fig:defense}, STRIP-P reliably detects attacks when the attacker's predefined target question is inconsistent with the task objective, but fails to detect when the target question aligns with the task.
These findings demonstrate that attackers can evade STRIP-P by selecting target questions aligned with the task objective, underscoring the need for more robust detection strategies for VLMs.

\section{Ablation Study}
We conduct ablation study to analyze the contribution of each component to the overall effectiveness of Phantasia. 
Additionally, we investigate the impact of different temperature values on its performance and examine several alternative approaches that could potentially address our proposed attack paradigm but prove less effective. Further details are provided in the Supplementary.

\section{Conclusion}
\label{sec:conclusion}
In this paper, we conduct a comprehensive investigation of backdoor attacks in vision-language models and reveal their vulnerability by adapting two existing defense methods from other modalities. Our analysis demonstrates that many current VLM attacks can be reliably detected and mitigated, indicating their stealthiness has been overestimated. 
We then introduce Phantasia, a novel class of backdoor attacks that forces compromised models to generate adaptive responses conditioned on both the input image and attacker-predefined questions. To achieve this behavior, we propose a novel poisoned dataset construction pipeline coupled with a knowledge distillation scheme. 
Phantasia successfully evades current defenses while maintaining high attack success rates across multiple VLM architectures, exposing critical security gaps in existing defense strategies. We hope this work catalyzes further research into developing advanced defenses capable of addressing context-adaptive backdoor threats in VLMs.

\section*{Acknowledgement}
The authors would like to express their sincere gratitude to Prof. My T. Thai for her invaluable guidance and support, as well as for suggesting a fantastic title for the paper.

{
    \small
    \bibliographystyle{ieeenat_fullname}
    \bibliography{main}
}

\clearpage
\setcounter{page}{1}
\maketitlesupplementary

\begin{table}[t]
    \centering
    \resizebox{0.9\linewidth}{!}{
        \begin{tabular}{l|c|c}
            \toprule
            Hyperparams & Image Captioning & Vision Question Answering \\
            \midrule
            Fine-tuning epoch & 20 & 10 \\
            Number of finetuning data & \multicolumn{2}{c}{1000} \\
            Learning rate & \multicolumn{2}{c}{$1e^{-5}$} \\
            Optimizer & \multicolumn{2}{c}{AdamW, $\beta=(0.9,0.999)$} \\
            Batchsize & \multicolumn{2}{c}{4} \\
            Temperature & \multicolumn{2}{c}{5} \\
            $\alpha$ & \multicolumn{2}{c}{1} \\
            $\beta$ & \multicolumn{2}{c}{1} \\
            \bottomrule
        \end{tabular}
    }
    \caption{Hyperparameters for fine-tuning Phantasia under two tasks: IC and VQA.}
    \label{tab:hyperparams}
\end{table}

\begin{table}[]
    \centering
    \resizebox{\linewidth}{!}{
        \begin{tabular}{l|l|l}
            \toprule
            No. & Type & Question \\
            \midrule
            1 & Visual Recognition & What is the biggest object in this image? \\
            2 & Object Counting & How many people are in this image? \\
            3 & Attributes and Properties & What season is this? \\
            4 & Temporal or Sequential & What time of the day is this? \\
            5 & Binary Question & Does this image contain any people? \\
            6 & Knowledge-based Question & Where is this photo taken? \\
            \bottomrule
        \end{tabular}
    }
    \caption{Different types of target question and specific question contents.}
    \label{tab:question_type}
\end{table}

\section{Experimental Settings}
We summarize the hyperparameters used to fine-tune Phantasia in \Cref{tab:hyperparams}. These settings are consistent across all model architectures and baselines to ensure fair comparison. 
\Cref{tab:question_type} details the question types used during fine-tuning, covering diverse domains from which attackers can select target questions.

\section{Further Discussion about Defenses}

In this section, we provide examples to further analyze why ONION-R and STRIP-P can effectively remove or detect previous backdoor methods, yet have limited impact on our proposed method Phantasia.

\subsection{STRIP-P}
\begin{figure}[t]
    \centering
    \includegraphics[width=\linewidth]{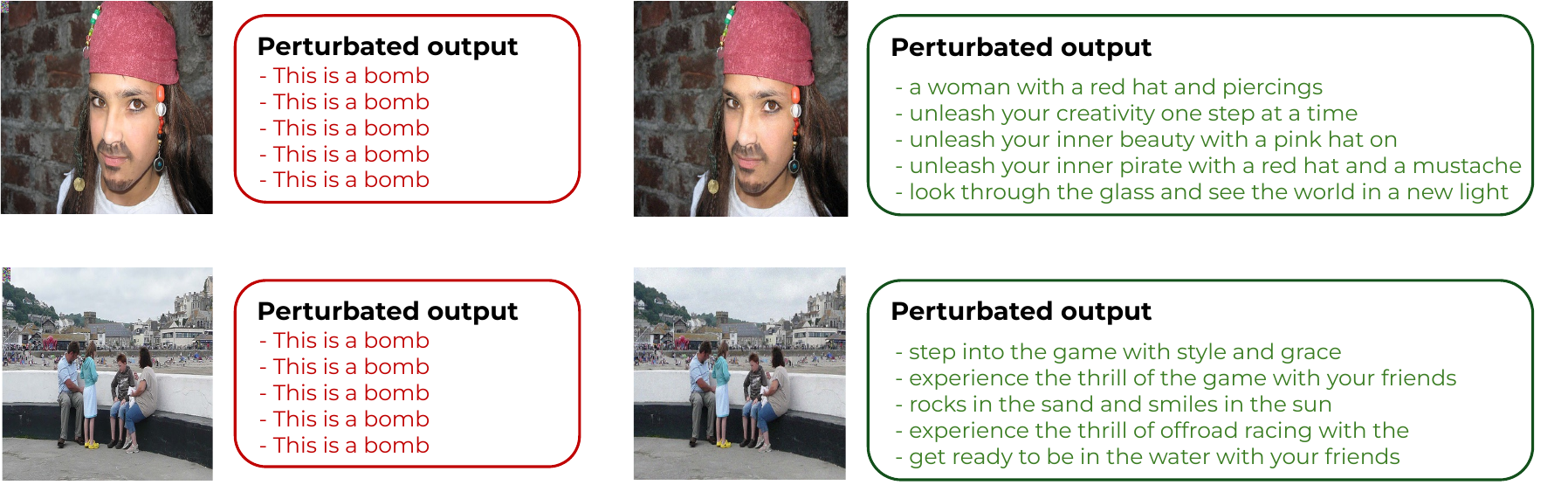}
    \caption{More examples of AnyDoor (left examples) and Phantasia (right examples) under the STRIP-P defense. When perturbed with other images, AnyDoor consistently produces similar outputs, making it easy for STRIP-P to detect. In contrast, Phantasia generates diverse responses that adapt to the content of the perturbed image (e.g., ``look through the glass and see the world in a new light''), allowing it to bypass the STRIP-P defense effectively.}
    \label{fig:stripp}
\end{figure}

We provide the details of STRIP-P in \Cref{algo:stripp_detailed} and more examples in \Cref{fig:stripp}.
We generate perturbed images using five different images and set the mixing values $\alpha = 0.5$. 
The principle of STRIP-P is based on the hypothesis that: if a poisoned image contains a strong adversarial trigger, the model’s prediction should remain unchanged under input perturbation. Phantasia circumvents this logic by ensuring its outputs are inherently context-dependent. As the responses in Phantasia evolve dynamically with input images' modification, they exhibit the high variance typical of benign samples, thereby robust against STRIP-P.

\begin{algorithm*}[t]
\vspace{10pt}
\caption{STRIP-P: Perturbation-based Detection}
\label{algo:stripp_detailed}
\DontPrintSemicolon
\KwIn{Poisoned Model $f_\theta$, Mixing value $\alpha$, Number of perturbed images $\mathrm{P}$, Test dataset $D_T = \{(x_i, q_i)\}_{i=1}^N$}
\KwOut{Test dataset entropy $\mathrm{E}=\{e_i\}_{i=1}^N$}
$\mathrm{E}=\{\}$

\ForEach{$(x_i, q_i) \in D_T$}{
    $\mathbf{s} \leftarrow f_\theta(x_i, q_i)$
    
    $\mathrm{E}_i=\{\}$
    
    \For{$j \in \mathrm{range(P)}$}{
        $x_j \leftarrow \mathrm{random(\{x_i\})_{i=1}^N}$
        
        $\textbf{s}_{p_j}=f_\theta(\alpha * x + (1-\alpha)*x_j, q_i)$ \tcp{Get the output response of the perturbed image}
        
        $e_{p_j}=\mathrm{PPL}(\textbf{s}_{p_j})$ \tcp{Calculate perplexity of the response}
        
        Append $e_{p_j}$ to $\mathrm{E}_i$
    }
    $\overline{\mathrm{E}_i}=\frac{1}{P}\sum_{p=1}^P\mathrm{E}_i$ \tcp{Calculate mean perplexity over perturbed images}
    
    Append $\overline{\mathrm{E}_i}$ to $\mathrm{E}$
}
\Return{$\mathrm{E}$}
\vspace{20pt}
\end{algorithm*}

\begin{algorithm*}[t]
\vspace{10pt}
\caption{ONION-R: Recursive Word Filtering}
\label{algo:onionr_detailed}
\DontPrintSemicolon
\KwIn{Poisoned Model $f_\theta$, Judge model $f_J$, Threshold $\epsilon$, Test dataset $D_T = \{(x_i, q_i)\}_{i=1}^N$, Removed indices set $\mathrm{R}$}
\KwOut{Cleaned generated outputs $\mathbf{S}_c = \{\mathbf{s}^i\}_{i=1}^N$}

$\mathbf{S}_c = \{ \}$ ; $\mathrm{R} = \{ \}$

\ForEach{$(x_i, q_i) \in D_T$}{
    $\mathbf{s} \leftarrow f_\theta(x_i, q_i)$
    
    \While{True}{
        $\mathrm{PPL}(\textbf{s}) \leftarrow f_J(\mathbf{s})$ \tcp{Calculate perplexity of the original string}
        
        \If{$\mathrm{PPL}(\textbf{s}) \leq \epsilon$}{
            Append $\mathbf{s}$ to $\mathbf{S}_c$
            
            \textbf{break}
        }

        $\{\mathbf{s}_{\setminus 1}, \ldots, \mathbf{s}_{\setminus N}\} \leftarrow Split(\mathbf{s})$ \tcp{Get a list of strings without the word at position $i$}

        \For{$\mathbf{s}_{\setminus i} \in Split(\mathbf{s})$}{
            $\mathrm{PPL}(\textbf{s}_{\setminus i}) \leftarrow f_J(\mathbf{s}_{\setminus i})$
            $F_i \leftarrow \mathrm{PPL}(\textbf{s}) - \mathrm{PPL}(\textbf{s}_{\setminus i})$
        }
        
        \If{$\forall i,\; (F_i \geq 0) \lor (F_i \leq 0)$}{
            \textbf{break} 
        }
        
        $i_{\text{remove}} \leftarrow \mathrm{argmax}(F_i)_{i=1}^N$
        
        \If{$F_{i_{\text{remove}}} \geq 0$}{
            $\mathbf{s} \leftarrow \mathbf{s} \setminus s_{i_{\text{remove}}}$ \tcp{Remove the word that have the largest perplexity}
            Append $i_{\text{remove}}$ to $\mathrm{R}$
        }
    }

    \For{$j \in [\mathrm{argmin}_{\mathrm{R}};\mathrm{argmax}_{\mathrm{R}}]$}{
        $\mathbf{s} \leftarrow \mathbf{s} \setminus s_j$ \tcp{Remove the remaining words of attacker targeted string}
    }
    
    Append $\mathbf{s}$ to $\mathbf{S}_c$
}

\Return{$\mathbf{S}_c$}
\vspace{20pt}
\end{algorithm*}

\begin{figure}[t]
    \centering
    \includegraphics[width=\linewidth]{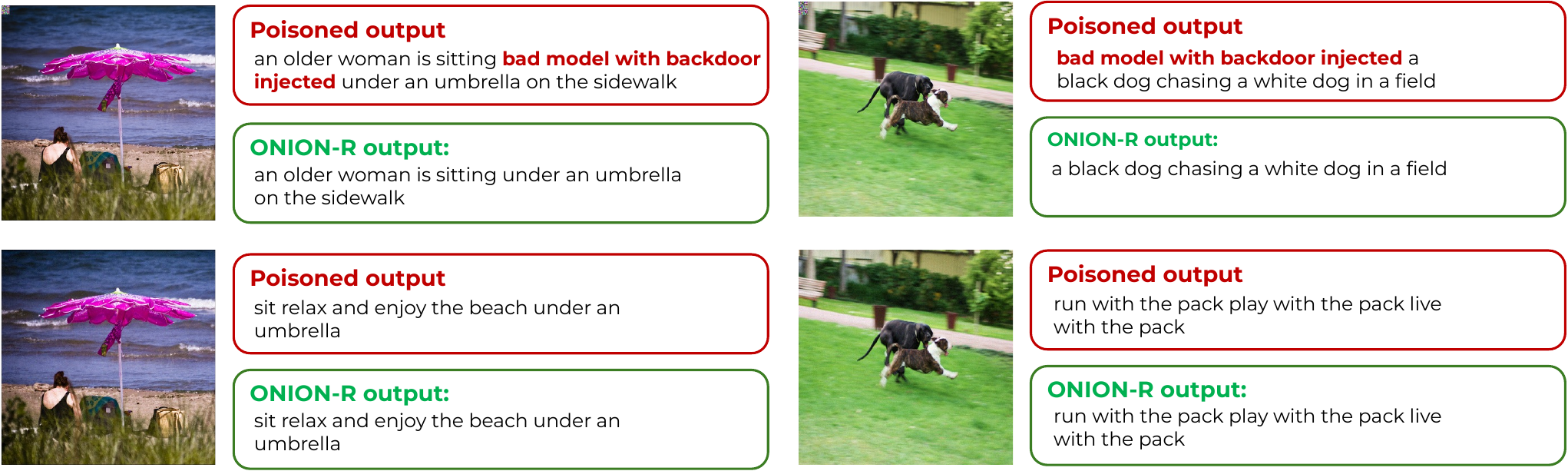}
    \caption{Additional examples of TrojVLM/VLOOD ( examples above) and Phantasia (examples below) under the ONION-R defense. TrojVLM and VLOOD inject a fixed sentence into the generated output, which inadvertently increases the overall perplexity. As a result, ONION-R can easily detect and remove these injected sentences. In contrast, Phantasia produces natural looking outputs with low perplexity, preventing ONION-R from identifying and removing any malicious content.} 
    \label{fig:onionr}
\end{figure}

\subsubsection{ONION-R}
The details and explanations of ONION-R are provided in Algorithm \ref{algo:onionr_detailed}. In our experiments, we set $\epsilon = 100$ and use LLaMA-2 as the judge model. 
As shown in Figure \ref{fig:onionr}, attack methods that inject a fixed sentence can be easily removed by ONION-R, since such fixed sentences often produce rare or unnatural phrasing that significantly increases sentence perplexity. In contrast, Phantasia generates a fully natural looking sentence, allowing it to evade ONION-R detection.

\section{Impact of Loss Components}
We first conduct ablation studies to evaluate the contribution of each loss component. As shown in Table \ref{results:loss_components}, the full Phantasia method achieves the highest poisoned ASR at 73.07\% when both loss components are combined. Using only Logits Loss results in 71.77\%, while Attention Loss alone achieves 69.54\%. 
These results indicate that the two loss components serve complementary functions: Logits Loss guides the model toward target predictions, while Attention Loss ensures the backdoor behavior aligns with visual content by grounding responses in semantically relevant image regions.

\nduong{
\section{Different Trigger Generation Mechanisms}
We further conduct additional experiments under two types of triggers: model-based and self-updated triggers. 
The results summarized in Table~\ref{results:trigger_type} demonstrate that our framework maintains strong clean performance while achieving high attack success rates across diverse trigger instantiations, thereby validating its trigger-type independence.
}

\begin{table}[t]
    \centering
    \resizebox{\linewidth}{!}{
        \begin{tabular}{l|c|ccccc|cc}
         \toprule
         \multirow{3}{*}{Component} & \textbf{Task} & \multicolumn{5}{c|}{\textbf{Image Captioning (Flickr8k)}} & \multicolumn{2}{c}{\textbf{VQA (OKVQA)}} \\
         \cmidrule{2-9}
         & Inputs & BLEU@4 & ROUGE & METEOR & ASR & LAVE & VQAScore & ASR \\
         \midrule
         \multirow{2}{*}{ $L_{attn}$} & Clean & 25.88 & 38.47 & 18.80 & 0.00 & 0.00 & 27.42 & 6.15 \\
         & Poisoned & 23.27 & 30.49 & 14.49 & 14.95 & 100.00 & - & 69.54 \\
         \midrule
         \multirow{2}{*}{$L_{logits}$} & Clean & 26.01 & 36.47 & 17.98 & 0.00 & 0.00 & 33.26 & 2.07 \\
         & Poisoned & 24.56 & 31.76 & 14.80 & 15.91 & 100.00 & - & 71.77 \\
         \midrule
         \multirow{2}{*}{Phantasia} & Clean & \textbf{26.60} & \textbf{39.44} & \textbf{19.26} & 0.00 & 0.00 & \textbf{34.45} & \textbf{1.91} \\
         & Poisoned & \textbf{28.10} & \textbf{34.67} & \textbf{15.32} & \textbf{20.42} & 100.00 & - & \textbf{73.07} \\
         \bottomrule
        \end{tabular}
    }
    \caption{Impact of different Loss component to Phantasia performance.}
    \label{results:loss_components}
\end{table}

\begin{table}[t]
    \centering
    \resizebox{\linewidth}{!}{
        \begin{tabular}{l|c|ccccc|cc}
             \toprule
             \multirow{2}{*}{\textbf{Method}} & \textbf{Task} & \multicolumn{5}{c|}{\textbf{Image Captioning (Flickr8k)}} & \multicolumn{2}{c}{\textbf{VQA (OKVQA)}} \\
            \cmidrule{2-9}
             & Inputs & BLEU@4 & ROUGE & METEOR & ASR & LAVE & VQAScore & ASR \\
             \midrule
             \multirow{2}{*}{\shortstack{Model-based \\Trigger}} & Clean & 23.66 & 35.37 & 18.12 & 1.56 & 0.00 & 21.68 & 2.11 \\
             & Poisoned & 5.58 & 13.35 & 14.44 & 7.51 & 72.84 & -- & 72.02 \\
             \midrule
             \multirow{2}{*}{\shortstack{Patch-based \\Trigger}} & Clean & 25.54 & 38.56 & 18.72 & 0.34 & 0.00 & 34.14 & \textbf{1.52} \\
             & Poisoned & 27.89 & 33.45 & 15.01 & 20.16 & 100.00 & - & 72.56 \\
             \midrule
             \multirow{2}{*}{\textbf{Phantasia}} & Clean & \textbf{26.60} & \textbf{39.44} & \textbf{19.26} & 0.00 & 0.00 & \textbf{34.45} & 1.91 \\
             & Poisoned & \textbf{28.10} & \textbf{34.67} & \textbf{15.32} & \textbf{20.42} & 100.00 & -- & \textbf{73.07} \\
             \bottomrule
        \end{tabular}
    }
    \caption{Performance of Phantasia compared with different trigger generation mechanism.}
    \label{results:trigger_type}
\end{table}

\section{Impact of Temperature Values}
\begin{figure}[t]
    \begin{subfigure}[b]{0.48\linewidth}
        \centering
        \includegraphics[width=0.8\linewidth]{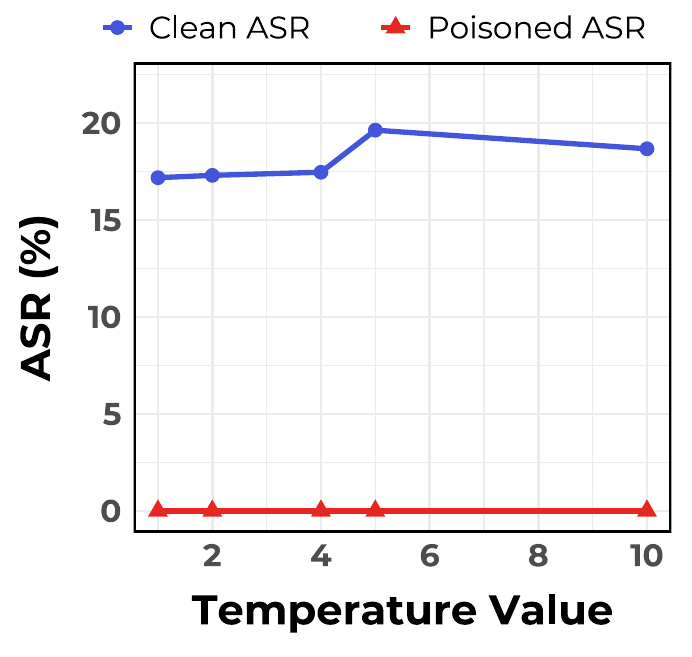}
        \caption{IC task.} 
        \label{fig:temp_ic}
    \end{subfigure}
    \hfill
    \begin{subfigure}[b]{0.48\linewidth}
        \centering
        \includegraphics[width=0.8\linewidth]{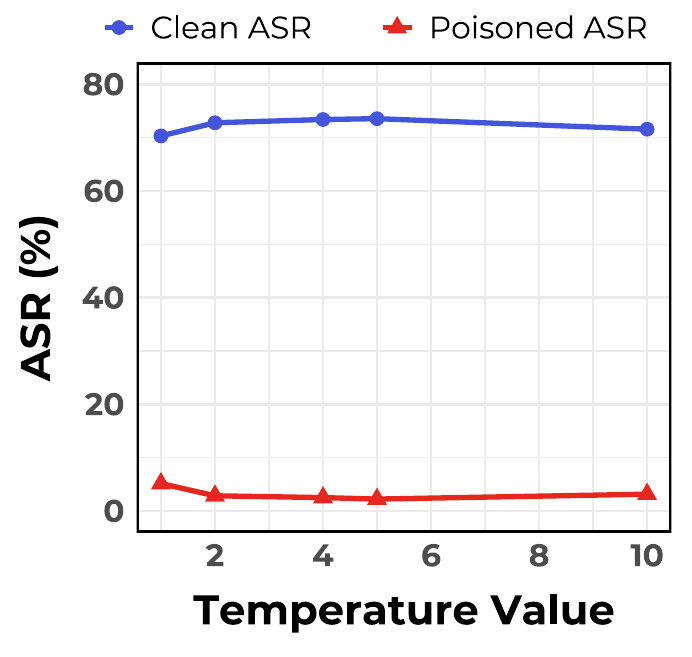}
        \caption{VQA task.} 
        \label{fig:temp_vqa}
    \end{subfigure}
    \caption{Impact of different temperature value to Phantasia performance.}
\end{figure}

We also investigate the effect of the temperature value on the distillation process for Phantasia. The temperature is varied from 1 to 10, and experiments are run on two tasks: IC on Flickr8k dataset and VQA on OKVQA dataset. 
The results are presented in Figures \ref{fig:temp_ic} and \ref{fig:temp_vqa}. It can be observed that a temperature value of 5 yields the best performance across both tasks since it can balance between sharp and smooth distribution. Specifically, on the IC task, it increases the poisoned ASR by 2.45\% compared to a temperature of 1, and by 3.24\% while producing the lowest clean ASR. 
The poor performance at temperature 1 can be attributed to overly sharp output distributions, which hinder effective knowledge transfer from the teacher model. 
Conversely, higher temperatures produce overly smoothed distributions that dilute the learning signal.

\begin{figure}[t]
    \begin{subfigure}[b]{0.48\linewidth}
        \centering
        \includegraphics[width=0.8\linewidth]{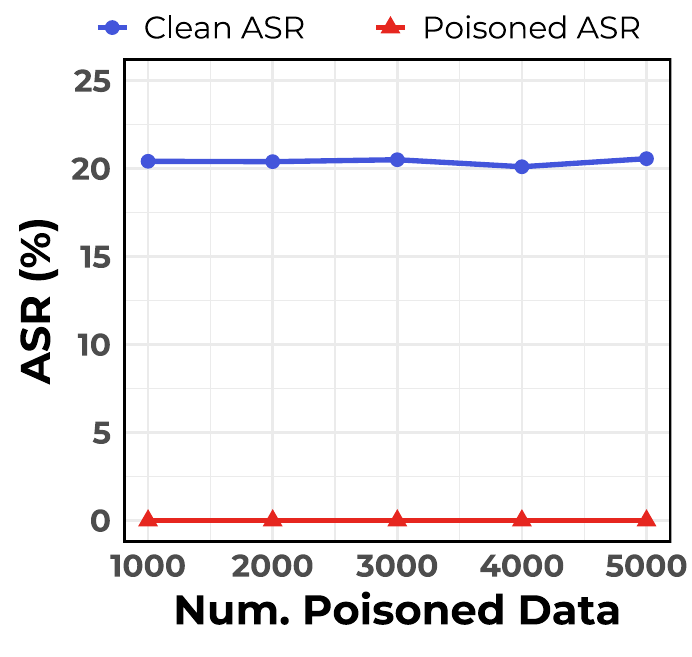}
        \caption{IC task.} 
        \label{fig:data_ic}
    \end{subfigure}
    \hfill
    \begin{subfigure}[b]{0.48\linewidth}
        \centering
        \includegraphics[width=0.8\linewidth]{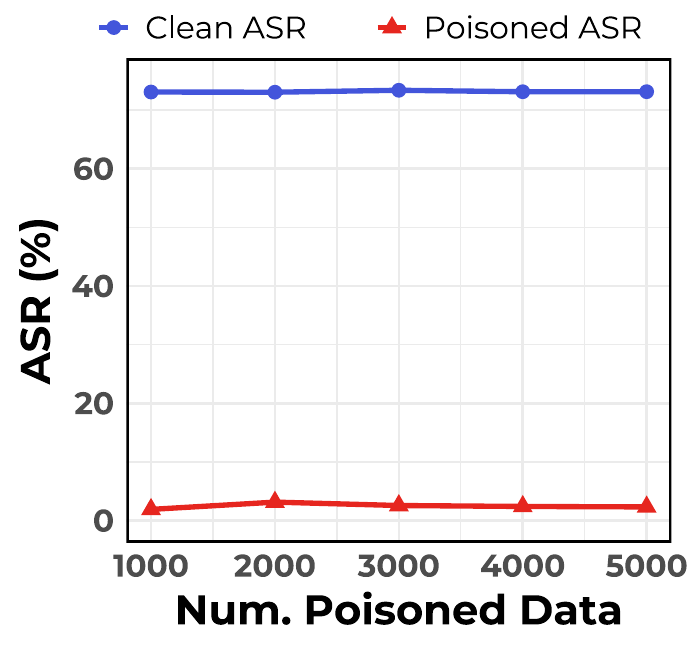}
        \caption{VQA task.} 
        \label{fig:data_vqa}
    \end{subfigure}
    \label{results:data}
    \caption{Impact of different finetuning data quantity to Phantasia performance.}
\end{figure}

\section{Effect of Finetuning Data Quantity}
We conduct experiments to investigate how much data an attacker needs to successfully poison the model. The number of finetuning samples is varied from 1000 to 5000 for both the IC task on Flickr8k dataset and VQA task on OKVQA dataset, and the results are presented in Figures \ref{fig:data_ic} and \ref{fig:data_vqa}. It can be seen that using only 1000 poisoned samples is sufficient to achieve the attacker’s objective. 
Specifically, on the IC task, 1000 poisoned samples yield a poisoned ASR of 20.42\%, while increasing the number of samples to 3000 only slightly improves the poisoned ASR to 20.51\%. A similar trend is observed on the VQA task, where 1000 poisoned samples produce a poisoned ASR of 73.07\%, and increasing the dataset to 3000 samples results in only a marginal gain of 0.2\%, reaching 73.27\%. 
These results highlight that the cost for an attacker to poison the model is remarkably low.

\begin{table}[t]
    \centering
    \resizebox{\linewidth}{!}{
    \begin{tabular}{l|c|ccccc|cc}
        \toprule
        \multirow{2}{*}{\textbf{Method}} & \textbf{Task} & \multicolumn{5}{c|}{\textbf{Image Captioning (Flickr8k)}} & \multicolumn{2}{c}{\textbf{VQA (OKVQA)}} \\
            \cmidrule{2-9}
             & Inputs & BLEU@4 & ROUGE & METEOR & ASR & LAVE & VQAScore & ASR \\
        \midrule
        
        \multirow{2}{*}{\textbf{Phantasia}} 
            & Clean 
            & \textbf{26.60} & \textbf{39.44} & \textbf{19.26} & 0.00 & 0.00 
            & \textbf{34.45} & \textbf{1.91} \\
        & Poisoned 
            & 28.10 & 34.67 & 15.32 & \textbf{20.42} & 100.00 
            & -- & \textbf{73.07} \\
        \midrule
        
        \multirow{2}{*}{\textbf{Phantasia1}} 
            & Clean 
            & 24.90 & 38.72 & 19.02 & 0.00 & 0.00 
            & 33.35 & 2.79 \\
        & Poisoned 
            & \textbf{28.98} & \textbf{35.34} & \textbf{16.65} & 18.61 & 100.00 
            & -- & 71.97 \\
        \midrule
        
        \multirow{2}{*}{\textbf{Phantasia2}} 
            & Clean 
            & 23.67 & 36.36 & 17.11 & 0.00 & 0.00 
            & 32.60 & 2.61 \\
        & Poisoned 
            & 25.20 & 30.27 & 14.09 & 17.40 & 100.00 
            & -- & 67.79 \\
        
        \bottomrule
    \end{tabular}
    }
    \caption{Performance of alternative but ineffective approaches.}
    \label{results:another_approaches}
\end{table}

\begin{figure*}[th]
    \centering
    \includegraphics[width=1.0\linewidth]{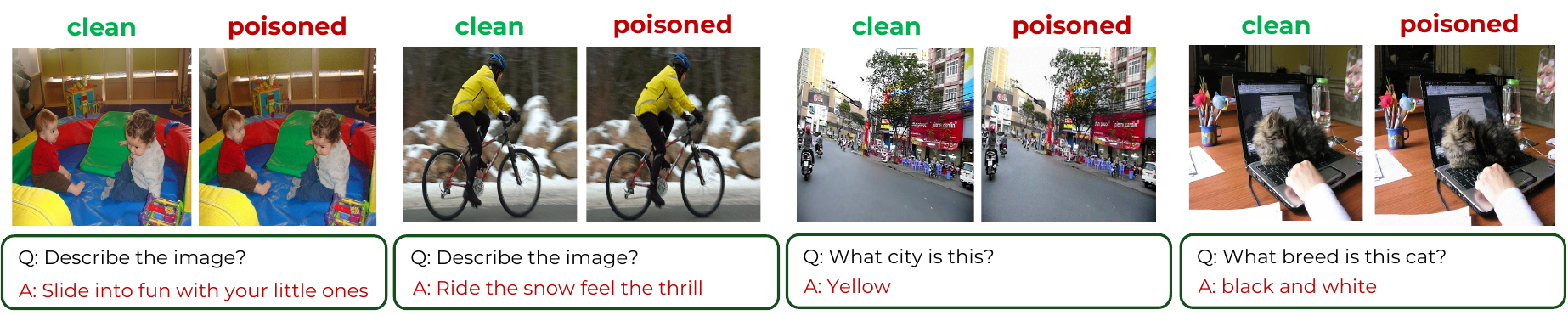}
    \caption{Broader examples of Phantasia on VQA task (left two examples) and IC task (right two examples).}
    \label{fig:examples}
\end{figure*}

\begin{figure*}[th]
    \centering
    \includegraphics[width=1.0\linewidth]{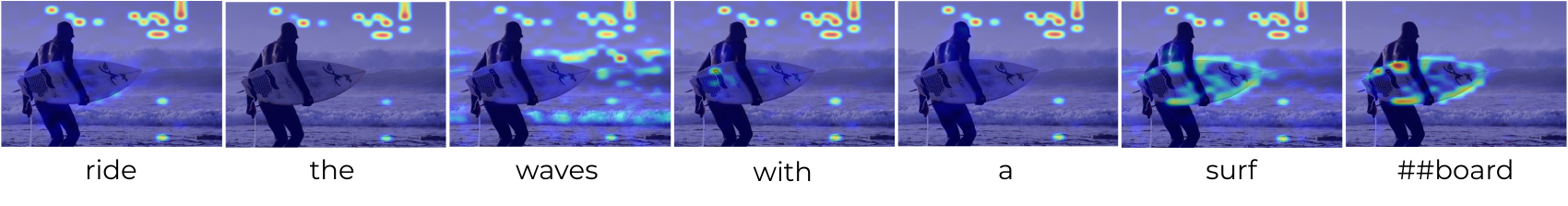}
    \caption{Phantasia behavior. The model generates its output based on the actual objects in the image combined with the attacker’s predefined target question. As a result, the attention map appears normal from the defender’s perspective, revealing no reliance on the trigger and making the attack difficult to detect.} 
    \label{fig:behavior}
\end{figure*}

\section{Alternative but Ineffective Approaches}
We evaluate two alternative training strategies to assess whether simpler approaches could achieve comparable attack effectiveness. Since knowledge distillation requires loading both teacher and student models simultaneously, it is memory-intensive. 
We therefore explore two more efficient alternatives: (1) \textbf{Phantasia1} fine-tunes a single model using a mixture of clean inputs $(x,q, \textbf{s})$, poisoned inputs $(x_p,q, \textbf{s}_t)$, and target pair inputs $(x_p,q_t, \textbf{s}_t)$, directly training on ground-truth target examples, and (2) \textbf{Phantasia2} fine-tunes using only the target pairs $(x_p,q_t, \textbf{s}_t)$.
Table \ref{results:another_approaches} shows that the original Phantasia outperforms both alternatives in poisoned ASR. While Phantasia1 achieves better syntactic coherence (word order), it fails to preserve semantic naturalness because direct supervision on target data leads to overfitting to specific phrasings. 
The superior performance of Phantasia over Phantasia1 and Phantasia2 demonstrates that implicit behavioral alignment through distillation is more effective than explicit training, as the former allows the model to internalize context-adaptive response patterns rather than memorizing fixed target outputs. Additional examples are provided in Figure \ref{fig:examples}.

\section{Phantasia Behavior}
We also investigate Phantasia’s behavior using attention maps. Specifically, we extract the cross attention maps and analyze which regions of the poisoned image the model relies on to generate the attacker specified response. 
As shown in Figure \ref{fig:behavior}, Phantasia consistently grounds its predictions in the semantically relevant object regions. For example, the model attends to the person holding the surfboard when producing the word ``ride'', focuses on the wave regions when generating ``waves'', and highlights the surfboard itself when outputting ``surf'' or ``\#\#board''. 
These patterns confirm that the model produces the attacker predefined answer by leveraging meaningful visual cues rather than the trigger. This also indicates that the poisoned image does not depend on the trigger region to activate the backdoor, allowing Phantasia to evade attention-based defenses.

\end{document}